\documentclass{article} 
\usepackage[final]{colm2026_conference}

\usepackage{microtype}
\usepackage{hyperref}
\usepackage{url}
\usepackage{booktabs}
\usepackage{xcolor}
\usepackage{amsmath}
\usepackage{amssymb} 
\usepackage{aas_macros}
\usepackage{caption}
\usepackage{array}
\usepackage{placeins} 
\usepackage[symbol]{footmisc}
\usepackage{graphicx}

\definecolor{darkblue}{rgb}{0, 0, 0.5}
\hypersetup{colorlinks=true, citecolor=darkblue, linkcolor=darkblue, urlcolor=darkblue}

\title{What AstroPT knows about galaxies,\\ and what that can teach us about LLMs}

\author{UniverseTBD$^*$: \\\textbf{Kshitij Duraphe, Aman Kumar$^{1}$, Michael J. Smith$^{2, 3}$, Shashwat Sourav$^{4}$}\\\\
$^1$IUCAA Pune, India, $^2$AstroAI, $^3$Center for Astrophysics $|$ Harvard \& Smithsonian \\ $^4$Washington University in St. Louis
}

\begin{document}

\maketitle
\begin{abstract}
Interpretability research increasingly asks \emph{when} concepts emerge during training and \emph{whether} linear probes recover real structure, but in language models these claims are hard to validate because language offers little ground-truth ordering of concepts or relationships among them. 
We propose the use of astronomical ground truth through \emph{AstroPT}, a transformer trained on millions of galaxy images, as a calibration testbed.
AstroPT is an LLM-like model trained within a domain where the difficulty ordering of concepts and the relations among them are known in advance. 
Probing frozen representations across checkpoints, layers, model sizes, and objective choices, we find that galaxy properties emerge in a fixed order that tracks their known difficulty---quantities written almost directly into the pixels (band magnitude) become decodable early in training and shallow in the network, while multiband/spectra based and inferred quantities (such as redshift and specific star formation rate) emerge later and deeper.
This order is invariant to our tested training objectives, and scales in magnitude but not in sequence with capacity. 
Our linear probe directions further recover the known physical structure among galaxy properties.
Our findings suggest that astronomy offers a controlled sandbox for calibrating mechanistic interpretability methods we otherwise apply to LLMs blind.
\end{abstract}

\footnotetext[1]{Alphabetical ordering, contributions at end of paper. Correspondence to: \texttt{mike@mjjsmith.com}.}

\section{Astronomy for machine learning}

Large language models are hard to interpret in part because their training data is broad, messy, and not grounded in known relations \citep{2001ApJ...550..212B,ref_bommasani2021}.
Scientific foundation models promise a cleaner setting.
In astronomy, many relationships in the data are already known from theory, observation, and long-standing empirical work \citep[e.g.][]{ref_courteau2007, ref_conroy2013, ref_universetbd2025, ref_dolag2025}.
This perspective permits us to ask two questions:
when such a model is trained, does its hidden space recover known scientific structure?
And can we use that structure as a tool to probe the model's learning dynamics and internalized knowledge?

Modern sky surveys contain millions of galaxy images tied to physical quantities such as brightness, size, stellar mass, morphology, redshift, and star formation history \citep{ref_mmu2024}.
These quantities form a natural ladder of difficulty.
Some are close to the raw pixels:
an \emph{apparent magnitude} is essentially the galaxy's integrated flux.
Other quantities require more physics knowledge and computation to recover---\emph{redshift} depends on multiple photometric bands or galaxy spectra, and the \emph{specific star formation rate} is more indirect still \citep{2007ApJ...660L..43N,2014ApJS..214...15S}.
Galaxy properties are also not independent: stellar mass scales with luminosity, star formation activity anti-correlates with stellar mass, and observed redshift correlates with mass due to survey selection effects.
This gives us both a difficulty ordering and a set of known relationships, so we can test whether a model merely internalizes only individual labels or also preserves the geometry that connects them.

\begin{figure}[htbp]
    \centering
    \includegraphics[width=\linewidth]{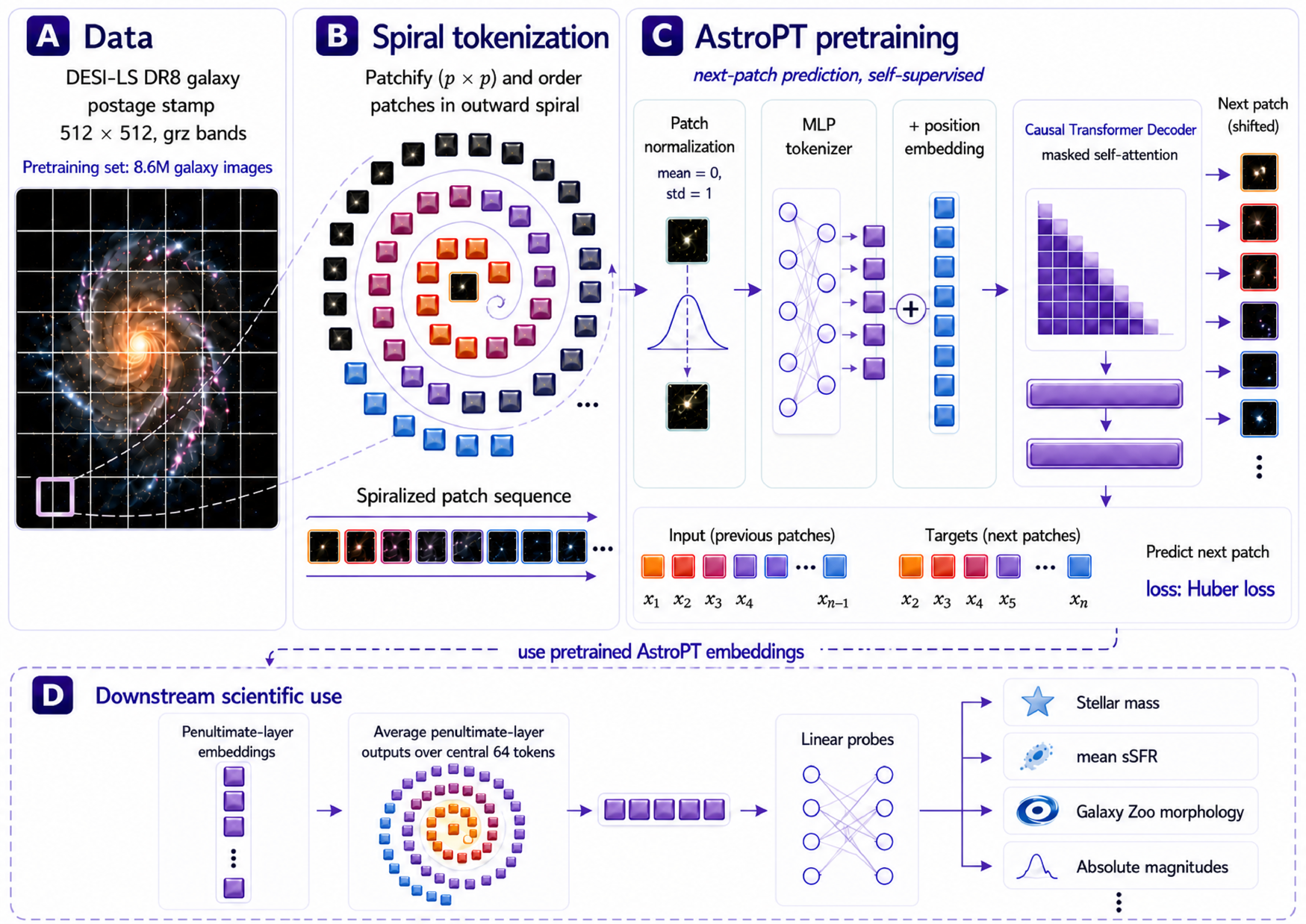}
    \caption{\textbf{Overview of AstroPT}. 
    Galaxy images are split into ordered patch sequences and used to train a GPT (pictured here) or BERT-style transformer via a patch reconstruction objective.
    The learned embeddings are then used for downstream probes of galaxy properties.}
    \label{fig_astropt}
\end{figure}

We study these questions using \emph{AstroPT} \citep[Fig.~\ref{fig_astropt};][]{ref_smith2024, ref_euclid2025}, a model deliberately built to resemble a large language model (LLM). 
It treats an astronomical observation as a sequence of data chunks: a galaxy image is split into ordered patches, the patches are treated like tokens, and the model learns in a self-supervised way by reconstructing dropped chunks in either an autoregressive GPT-like or masked autoencoding BERT-like fashion \citep{ref_radford2018,ref_devlin2018}. 
This architectural choice follows \citet{ref_sutton2019}: a scalable general architecture should win out over a bespoke one, and the precise design matters little provided it scales, so the most developed and community-supported option is preferred. 
That option---the causal autoregressive transformer---is also technically convenient for the many modal domain of astronomy, as it facilitates efficient training and modality fusion by simple token-chaining. 

Our goal is not to claim that AstroPT discovers new galaxy laws from scratch.
Instead, we ask whether our interpretability tools recover this known structure inside an LLM-like model, whether the order in which that structure emerges tracks known difficulty, whether the latent space geometry preserves the known relationships, and whether these signals survive changes of training recipe.
This turns astronomy into a controlled testbed for the developmental, probe-based interpretability methods we otherwise apply to LLMs blind. 

We find that: 
\textbf{(1)} The order in which galaxy properties become decodable during pre-training tracks their known separation from low-level pixel statistics, and the same order recurs along network depth.
We find that directly recoverable quantities emerge early in training and shallow in the network, more complex quantities later and deeper. 
\textbf{(2)} This order is invariant to training objective, and scales in magnitude but not in sequence with model capacity, so it is a property of the learning problem rather than of the recipe.
\textbf{(3)} Our physics probe directions recover the known sign structure among galaxy properties.

\section{Training and probing LLM-like models with astronomical data}\label{sec:setup}

AstroPT treats a galaxy image as a sequence of patches and is trained by patch reconstruction, the visual analog of token prediction \citep{ref_smith2024}. 
We train our AstroPT models with both GPT-like autoregressive (AR) \citep{ref_radford2018} and BERT-like masked-autoencoding (MAE) objectives \citep{ref_devlin2018}, 
Loosely following Pythia \citep{ref_biderman2023} we train across model sizes spanning  \{1M, 21M, 100M\} parameters.
We save checkpoints through one epoch on 8.7M galaxy postage stamps from the \texttt{Smith42/galaxies}\footnote{\url{https://hf.co/datasets/Smith42/galaxies}} DESI Legacy Survey dataset \citep{ref_dey2019,ref_walmsley2023}. 
For each frozen checkpoint we extract hidden states at every layer, and fit linear ridge probes for galaxy properties, reporting held-out $R^2$. 
A property being linearly decodable does not prove the model uses it during pre-training, but it shows the information is organized accessibly in the hidden state. 
Our targets form an \emph{a priori} difficulty ladder: quantities recoverable almost directly from the pixels ($\text{mag}\,r$; $r$-band magnitude), a wavelength-integrated quantity ($z$; redshift), and an inferred one ($\log \text{sSFR}$; specific star formation rate).

\begin{figure}[htbp]
    \centering
    \includegraphics[width=0.95\textwidth]{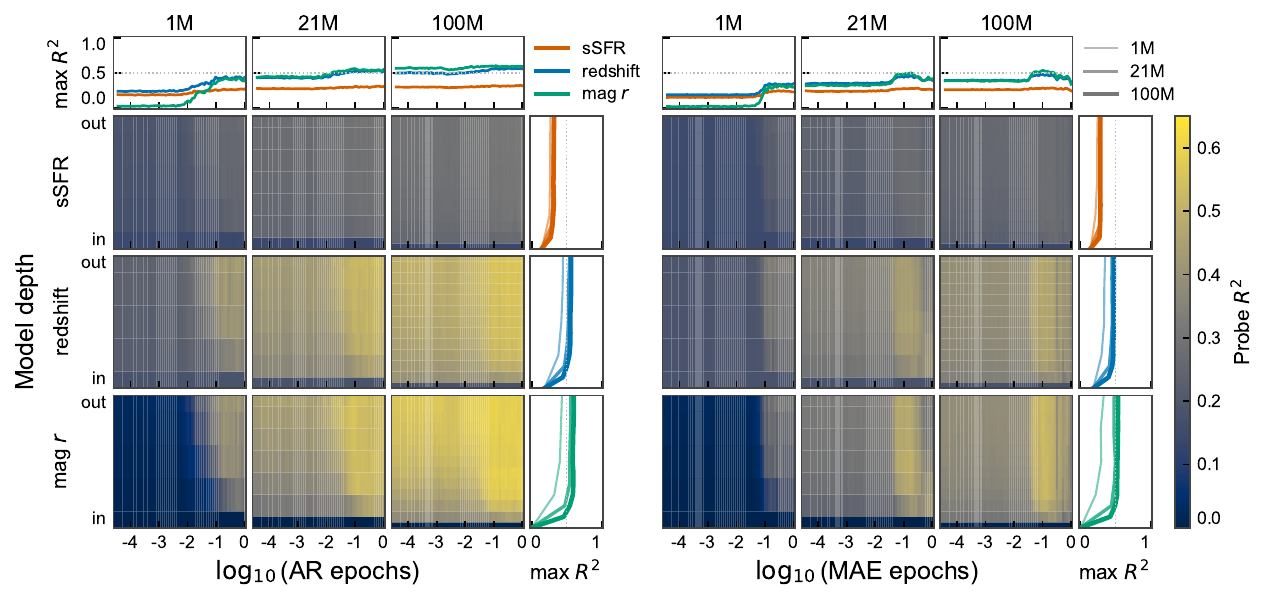}
    \caption{\textbf{Concepts emerge in a fixed, difficulty-ordered sequence.} Held-out $\text{mag}\,r$, redshift, and sSFR $R^2$ as a function of pre-training progress and model depth.
    Directly recoverable photometric information ($r$-band magnitude) is decoded earliest and most strongly; redshift follows more weakly; specific star formation rate remains weakly retrievable after one epoch at our model parameter counts. 
    Likewise, our directly recoverable quantities peak in shallow layers, while integrated and inferred quantities peak deeper.
    The order is fixed across model size, with capacity raising the attained score rather than reordering the properties.}
    \label{fig:physical_emergence}
\end{figure}

\textbf{More difficult properties emerge later in our training runs.}
The upper line plots in Fig.~\ref{fig:physical_emergence} show held-out $R^2$ as a function of training progress. 
We find that our probed properties do not become decodable together, but that the order in which they emerge tracks their known difficulty. 
$\text{mag}\,r$ rises first and highest, redshift follows, and sSFR stays weak and does not robustly emerge within one epoch.
The ordering is fixed across model size, with average probe performance rising with model capacity as expected.

\textbf{More difficult properties emerge later in our network architectures.}
The same `difficulty' ordering also appears as a function of network depth. 
The rightmost line plots in Fig.~\ref{fig:physical_emergence} show layer-wise held-out $R^2$.
$r$-band magnitude is decodable early and broadly across layers, while redshift is weak near the input and becomes stronger in later layers.
sSFR remains weak across depth, showing that it too complex to learn accurately in a single epoch at our tested model capacity.
Along with the training time curves, the heatmaps show us that easier photometric features appear early while more integrated physical quantities become accessible later in training and deeper in the network.

\textbf{The order of property emergence is not dependent on objective.}
If the emergence order were an artifact of the training objective rather than of the data, it should reorder when we change the objective. 
Fig.~\ref{fig:physical_emergence} shows it does not.
We see that the AR and MAE objectives reproduce the same difficulty ordering across our tested model capacities. 
The ordering therefore reflects what the representation must encode about galaxies, and is not a property of the model pre-training objective. 

\begin{figure}[htbp]
    \centering
    \includegraphics[width=0.95\linewidth]{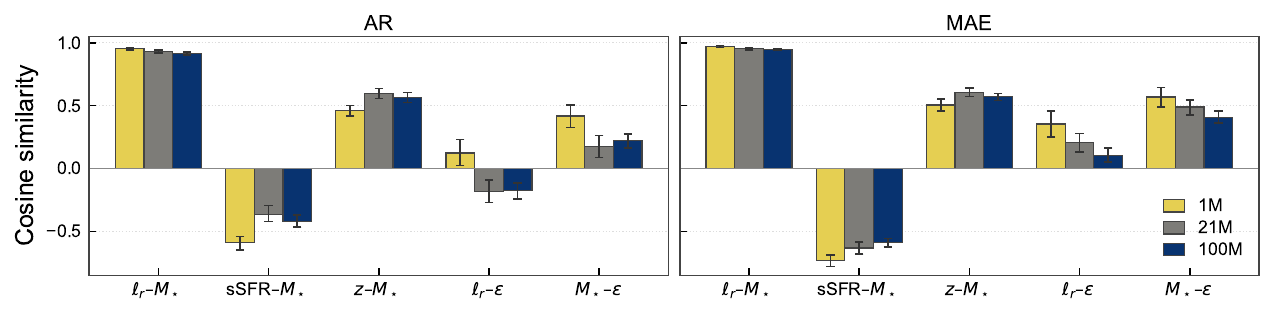}
    \caption{\textbf{Probe geometry recovers the signs of known galaxy relationships.} Across model sizes, luminosity and stellar-mass directions are strongly aligned, sSFR and stellar-mass directions are anti-aligned,  redshift and stellar-mass directions are positively aligned, and the mass residual is much less aligned with luminosity than stellar mass is, matching the known sign structure of these relationships.}
    \label{fig:relationship_geometry}
    \vspace{-0.5em}
\end{figure}

\textbf{Our probes' directions recover known physical structure.}
Beyond single properties, the directions that our probes assign to different quantities encode the relationships among them. Fig.~\ref{fig:relationship_geometry} shows that across our model sweep the luminosity and stellar-mass directions are strongly aligned, the sSFR and mass directions anti-aligned, and the redshift and mass directions positively aligned---the known sign structure of these relationships.
We can predict these from known galaxy physics.
Luminosity tracks stellar mass because more massive galaxies host more stars; within the star-forming population galaxies follow the well-established main sequence, $\log \text{SFR} = \alpha \log M_\star + \beta$, with a sub-linear slope $\alpha < 1$ \citep{2014ApJS..214...15S}, so since $\text{sSFR} \equiv \text{SFR}/M_\star$ the sSFR--mass direction is mildly anti-aligned; and the redshift--mass alignment arises primarily from a Malmquist-style selection effect \citep{ref_malmquist1922}, as fainter high-redshift galaxies enter a flux-limited sample only when they are brighter and more massive. 
We also ask a deeper question about the models' relational encoding of mass and luminosity.
To do so we first use $\epsilon$ to denote the mass residual at fixed luminosity, $\epsilon_{M|\ell_r}=\log M_\star-(a\ell_r+b)$, the part of stellar mass that luminosity alone cannot predict. 
We define the absolute $r$-band log-luminosity as $\ell_r=-0.4M_r$, where $M_r$ is the absolute magnitude.
A model treating mass as a mere luminosity shortcut would keep the residual direction aligned with the luminosity direction; instead the residual direction is nearly orthogonal to luminosity, increasingly so in our larger models, suggesting AstroPT represents mass-at-fixed-luminosity as a distinct factor. Label-space relationships and their  drivers, probe construction, and bootstrapped confidence intervals are shown in Apps.~\ref{app:redshift},~\ref{app:residual}.

\textbf{Limitations.} Galaxies are not language and image patches are not discrete tokens; we do not claim here that astronomical data is itself useful for the pre-training of LLMs, but that a model which learns like an LLM allows us to calibrate and explore interpretability methods we otherwise apply  blindly.
While the physical property emergence ordering we recover in this work is robust across our tested objectives and model sizes, several limitations bound its reach and mark avenues we will address in extended future work:
our analysis is correlational; our physical labels are catalog-inferred; pre-training is limited to one epoch and a relatively small sweep of \{1M, 21M, 100M\} parameters; and our linear probes reveal only what is linearly accessible in our tested embedding spaces.

\textbf{Towards an `answer key' for mechanistic interpretability.}
We find that galaxy properties become decodable in order of physical complexity across both the extent of pre-training run, and also along network depth.
We expect this is because features more directly recoverable from the galaxy image input should emerge earlier in training and shallower in the network, while features that require more complicated integration or inference should emerge later and deeper.
We find that the sequence of decodability is invariant to the training objective and to model size.
Our probe directions reproduce the known sign structure among physical properties, suggesting that the models internalise the properties in a uniform physically-coherent manner.
This is direct evidence that our probes read out real physical structure---evidence that cannot be easily replicated in language due to data constraints.
We have undertaken this brief case study as we believe that astronomy offers an underused and underappreciated mechanistic toolbox for the study of language models; astronomy enjoys an abundance of openly accessible data, well-defined physical relations that govern its data, and finds itself happily embroiled within a bubbling wave of foundation model development \citep{ref_smith2023}.
We close in the hope that this work encourages more research and effort in the young field of `astronomy for AI'.

\FloatBarrier

\section*{Contributions}

KD developed the embedding processing and extraction software, conducted investigations informing property selection  and tokenization strategy, and contributed to reviewing and editing the manuscript.

AK ran experiments before finalization of the experiment code, and reviewed and edited the manuscript.

MJS wrote the AstroPT pre-training code, created figures 2 and 3, wrote the body of the paper, contributed to review and editing, and guided project work.

SS designed which astronomy equations to use and discover, writing, reviewing the initial and final draft, making figure 1, table 1, doing inference and preparing the initial figure 2 and 3. Prepared figures 5--12. 

All authors conceptualized the project via `strong aura'.

\bibliography{colm2026_conference}
\bibliographystyle{colm2026_conference}

\appendix

\clearpage

\section{Linear probe training and alignment details}\label{app:redshift}

For each checkpoint and layer we  extract the representation for each galaxy image, and train linear ridge probes for several catalog quantities: $\text{mag}\,r$, an absolute $r$-band luminosity proxy, stellar mass, specific star formation rate, and redshift. We define the luminosity proxy as
$
\ell_r \equiv -0.4 M_r ,
$
where $M_r$ is the absolute $r$-band magnitude; the factor $-0.4$ follows from $M_r=-2.5\log_{10}L_r+C$, with the additive constant absorbed by the intercept of the linear fit. 
For a target $y$ the probe is $\hat{y}=h^\top w_y+b_y$, and we use the normalized weight vector $v_y=w_y/\|w_y\|$ as the model-space direction for that property, comparing directions by cosine similarity. Positive cosine means two properties are represented along similar directions, negative cosine means opposite directions, and values near zero mean the directions are orthogonal.

\subsection{Supporting relational-geometry plots}\label{app:label_space}

Before studying the model representation, we first check the relationships already present in the catalog labels, since our goal is to ask whether known relationships are preserved inside the model's hidden space rather than to rediscover them.

\begin{figure}[h]
    \centering
    \includegraphics[width=\linewidth]{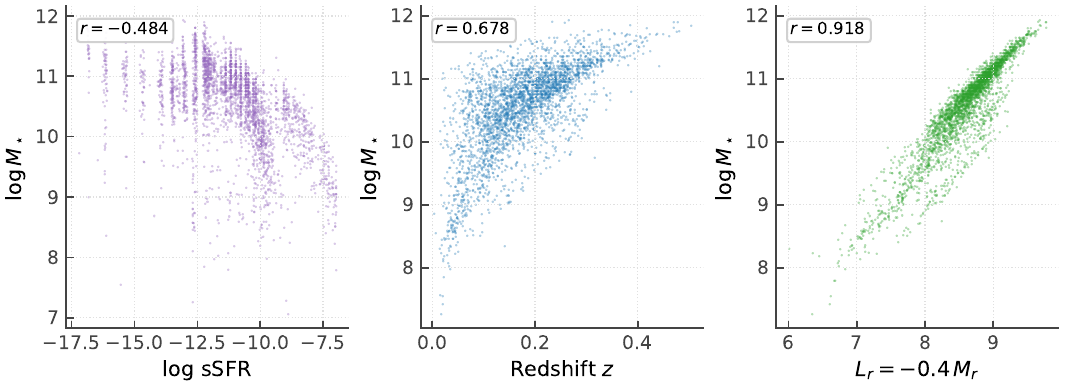}
    \caption{
    \textbf{Label-space galaxy relationships used as baselines for the representation-geometry analysis.}
    Left: specific star formation rate is anti-correlated with stellar mass.
    Middle: redshift is positively correlated with stellar mass.
    Right: absolute $r$-band luminosity is strongly correlated with stellar mass.
    These label-space relationships set the expected signs for the probe-direction analysis in the main text.
    }
    \label{fig:app_label_space}
\end{figure}

Fig.~\ref{fig:app_label_space} shows the raw catalog relationships, each with a known physical origin that fixes the sign we expect the probe directions to recover.
Luminosity and stellar mass are strongly related (right), as the presence of luminous stars means more mass.
Specific star formation rate is mildly anti-correlated with stellar mass (left).
This is because, within the star-forming population, galaxies follow the well-established main sequence ($\log \text{SFR} = \alpha \log M_\star + \beta$), with a sub-linear slope $\alpha < 1$ \citep{2014ApJS..214...15S}. 
Since $\text{sSFR} \equiv \text{SFR}/M_\star$ it follows that
\begin{equation*}
    \log \text{sSFR} \propto (\alpha - 1)\log M_\star, \quad \alpha < 1, 
\end{equation*}
a mildly negative trend. 
This is reinforced by the rising quiescent (non-star-forming) fraction at high stellar mass \citep{ref_wetzel2012}, though both effects are weak relative to the several orders of magnitude that sSFR spans at fixed mass.
Redshift and stellar mass are positively correlated (middle), arising primarily from a Malmquist-style selection effect \citep{ref_malmquist1922}.
Higher-redshift galaxies appear fainter, so only intrinsically luminous (i.e. more massive) galaxies exceed the survey detection threshold, biasing high-redshift samples toward larger stellar masses.
Higher-redshift galaxies are also observed to have higher star-formation rates \citep{2014ARA&A..52..415M}, which further contributes to the sSFR--redshift correlation.
In short, these label-space signs result from intrinsic galaxy physics and survey selection effects, and set the expectations for the probe-direction analysis in the main text.

\subsection{Training-time stability of relationship geometry}

\begin{figure}[htbp]
    \centering
    \includegraphics[width=0.8\linewidth]{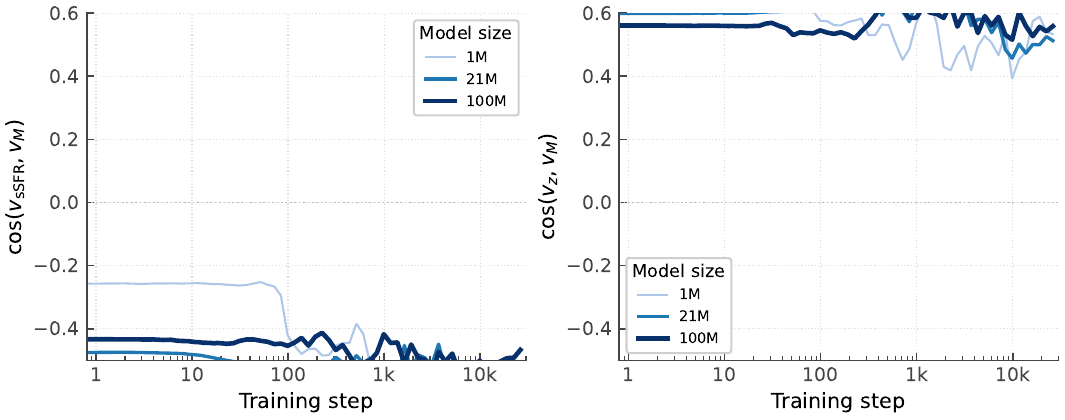}
    \caption{
    \textbf{Training-time relationship geometry under the AR objective.}
    Left: the sSFR direction is anti-aligned with the stellar-mass direction through training.
    Right: the redshift direction is positively aligned with the stellar-mass direction through training.
    The signs are also stable across model sizes, suggesting that the relationships are not confined to our fully pre-trained models.
    }
    \label{fig:app_geometry_emergence}
\end{figure}

\begin{figure}[htbp]
\centering
\includegraphics[width=0.8\linewidth]{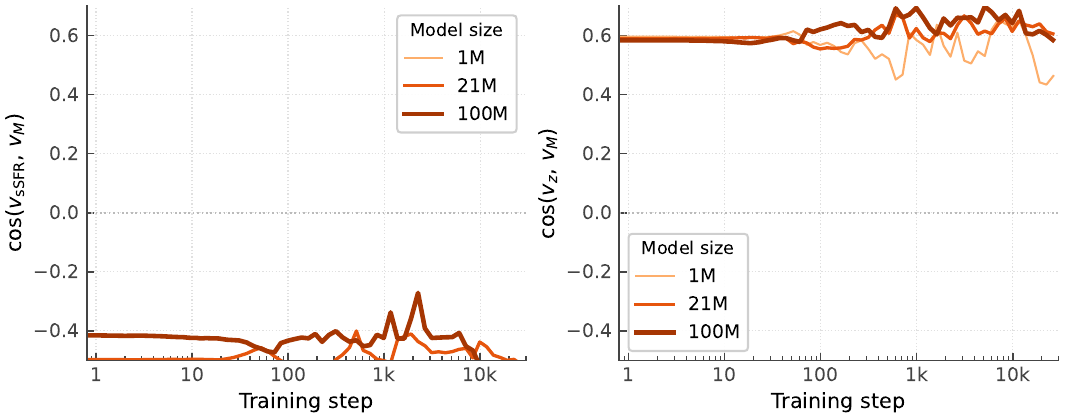}
\caption{
    \textbf{Training-time relationship geometry under the MAE objective.}
    Left: the sSFR direction is anti-aligned with the stellar-mass direction through training.
    Right: the redshift direction is positively aligned with the stellar-mass direction through training.
    The signs are also stable across model sizes, suggesting that the relationships are not confined to our fully pre-trained models.
}
\label{fig:app_geometry_mae_training}
\end{figure}

Figs.~\ref{fig:app_geometry_emergence} and \ref{fig:app_geometry_mae_training} show that the signs of the cosine physical property relationships are stable over our pre-training runs.

\subsection{Layer-wise relationship geometry}

\begin{figure}[htbp]
    \centering
    \includegraphics[width=0.8\linewidth]{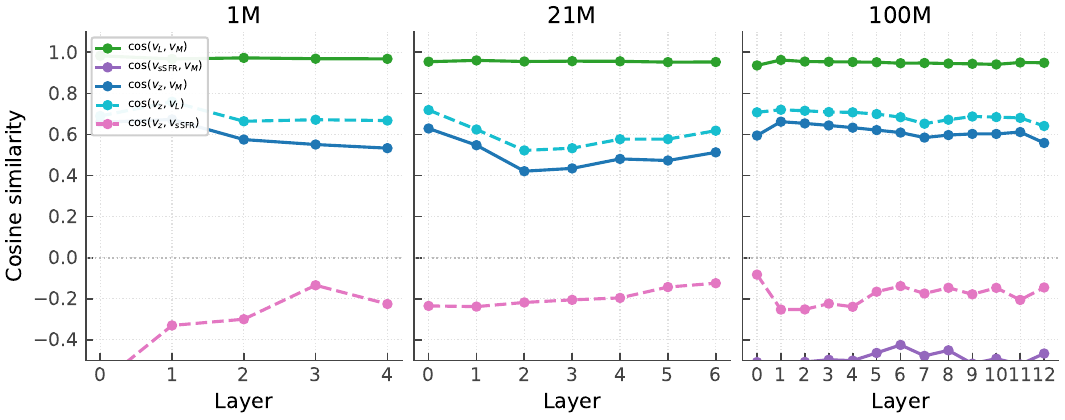}
    \caption{
    \textbf{Layer-wise relationship geometry under the AR objective.}
    The luminosity and stellar-mass directions remain strongly aligned across layers, the sSFR and stellar-mass directions remain anti-aligned, and the redshift direction is positively aligned with both mass and luminosity.
    This suggests the relationship geometry is distributed across layers rather than confined to the final layer.
    }
    \label{fig:app_geometry_layers}
\end{figure}

\begin{figure}[htbp]
\centering
\includegraphics[width=0.8\linewidth]{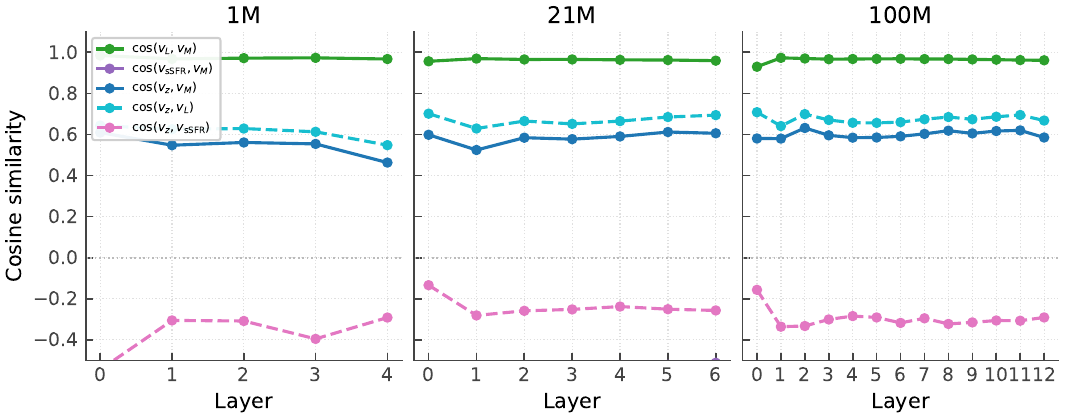}
\caption{
\textbf{Layer-wise relationship geometry under the MAE objective.}
Luminosity and stellar mass remain strongly aligned across layers, sSFR and stellar mass remain anti-aligned, and redshift remains positively aligned with both mass and luminosity. The same sign structure appears under MAE and AIM, supporting the claim that the recovered geometry is not tied to a single training objective.
}
\label{fig:app_geometry_mae_layers}
\end{figure}

Figs.~\ref{fig:app_geometry_layers} and \ref{fig:app_geometry_mae_layers} show the same qualitative relationships across layers, supporting the claim that AstroPT's hidden space preserves galaxy relationships across its' layers in both the AR and MAE regimes.

\subsection{Final-checkpoint bootstrap metrics}
\label{app:bootstrap}

\begin{table}[htbp]
\centering
\scriptsize
\setlength{\tabcolsep}{3pt}
\caption{
Final-checkpoint bootstrap metrics across objective and model-size choices.
Here $\ell_r=-0.4M_r$ is the absolute $r$-band log-luminosity proxy, $M_\star$ is stellar mass, and $\epsilon_{M|\ell_r}$ is the residual mass at fixed luminosity.
Intervals are 95\% bootstrap intervals.
}
\label{tab:bootstrap_metrics}

\resizebox{\linewidth}{!}{
\begin{tabular}{lccccc}
\toprule
Config & $R^2(\ell_r)$ & $R^2(M_\star)$ & $R^2(\epsilon_{M|\ell_r})$ & $R^2(\log{\rm sSFR})$ & $R^2(z)$ \\
\midrule
AR 1M    & 0.422 [0.420, 0.424] & 0.535 [0.533, 0.536] & 0.101 [0.099, 0.104] & 0.258 [0.256, 0.260] & 0.410 [0.408, 0.412] \\
AR 21M   & 0.499 [0.496, 0.501] & 0.610 [0.608, 0.612] & 0.109 [0.104, 0.114] & 0.298 [0.295, 0.302] & 0.503 [0.499, 0.506] \\
AR 100M  & 0.530 [0.526, 0.533] & 0.639 [0.636, 0.641] & 0.110 [0.103, 0.116] & 0.309 [0.304, 0.313] & 0.573 [0.568, 0.576] \\
\midrule
MAE 1M   & 0.356 [0.354, 0.358] & 0.454 [0.452, 0.456] & 0.085 [0.082, 0.087] & 0.248 [0.245, 0.250] & 0.328 [0.326, 0.330] \\
MAE 21M  & 0.350 [0.347, 0.354] & 0.446 [0.443, 0.449] & 0.070 [0.065, 0.075] & 0.228 [0.224, 0.232] & 0.323 [0.319, 0.327] \\
MAE 100M & 0.386 [0.381, 0.390] & 0.488 [0.484, 0.491] & 0.080 [0.075, 0.085] & 0.246 [0.243, 0.250] & 0.349 [0.343, 0.354] \\
\bottomrule
\end{tabular}
}

\vspace{1em}

\resizebox{\linewidth}{!}{
\begin{tabular}{lccccc}
\toprule
Config & $\cos(\ell_r,M_\star)$ & $\cos({\rm sSFR},M_\star)$ & $\cos(z,M_\star)$ & $\cos(\ell_r,\epsilon)$ & $\cos(M_\star,\epsilon)$ \\
\midrule
AR 1M    & 0.951 [0.940, 0.960] & -0.590 [-0.647, -0.540] & 0.462 [0.420, 0.502] & 0.123 [0.024, 0.231] & 0.418 [0.327, 0.503] \\
AR 21M   & 0.931 [0.919, 0.942] & -0.364 [-0.422, -0.295] & 0.599 [0.556, 0.636] & -0.180 [-0.270, -0.089] & 0.175 [0.089, 0.262] \\
AR 100M  & 0.915 [0.903, 0.925] & -0.420 [-0.466, -0.369] & 0.563 [0.524, 0.604] & -0.177 [-0.241, -0.112] & 0.222 [0.163, 0.276] \\
\midrule
MAE 1M   & 0.970 [0.961, 0.977] & -0.731 [-0.780, -0.687] & 0.504 [0.455, 0.551] & 0.355 [0.249, 0.459] & 0.568 [0.487, 0.643] \\
MAE 21M  & 0.951 [0.942, 0.959] & -0.632 [-0.680, -0.586] & 0.604 [0.572, 0.638] & 0.206 [0.133, 0.280] & 0.489 [0.424, 0.547] \\
MAE 100M & 0.946 [0.940, 0.953] & -0.589 [-0.622, -0.557] & 0.571 [0.541, 0.595] & 0.104 [0.051, 0.164] & 0.408 [0.363, 0.458] \\
\bottomrule
\end{tabular}
}
\end{table}

Tab.~\ref{tab:bootstrap_metrics} reports final-checkpoint bootstrap metrics across objective and model-size choices. Here we provide the final-checkpoint uncertainty estimates for the main probe quantities and the relational-geometry analysis. The $R^2$ block includes the luminosity proxy, stellar mass, the mass residual, sSFR, and redshift. The cosine block supports the probe-geometry result in Fig.~\ref{fig:relationship_geometry}. Across configurations, luminosity and stellar mass remain strongly aligned, sSFR and stellar mass remain anti-aligned, and redshift and stellar mass remain positively aligned.

\pagebreak

\section{Residual mass-at-fixed-luminosity and activation patching}\label{app:residual}

This appendix reports an exploratory analysis of AstroPT representations beyond the dominant luminosity-mass axis. 
For each checkpoint and layer we freeze AstroPT and extract the  representation $h$ for each galaxy image, then train the same probe directions defined in App.~\ref{app:redshift} (linear ridge probes for the luminosity proxy $\ell_r=-0.4M_r$, stellar mass, sSFR, and redshift, with directions compared by the cosine similarity of their normalized weight vectors $v_y=w_y/\|w_y\|$).

We test whether stellar mass is represented only through luminosity. We fit the empirical mass--luminosity relation on the training split, $\log M_\star = a\ell_r+b$, and define the mass residual at fixed luminosity \citep{1999astro.ph..5116H,2022arXiv220600989H},
\begin{equation*}
   \epsilon_{M|\ell_r}=\log M_\star-(a\ell_r+b),
\end{equation*}
which measures whether a galaxy is more or less massive than expected from its $r$-band luminosity alone, reflecting variation in mass-to-light structure (stellar-population age, star formation history, dust, metallicity, IMF, or catalog-model assumptions).

\begin{figure}[htbp]
   \centering
    \includegraphics[width=0.6\linewidth]{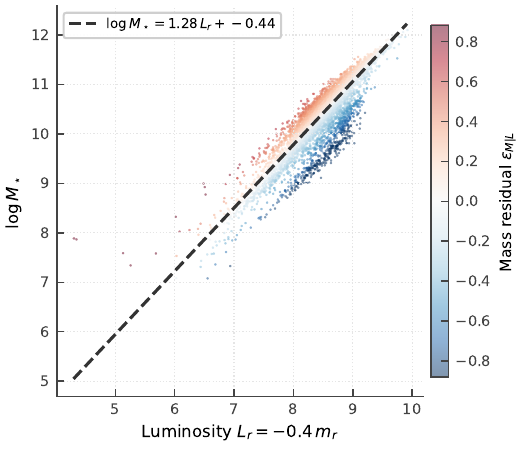}
   \caption{
   \textbf{Empirical mass--luminosity relation used to define the residual mass at fixed luminosity.}
   The horizontal axis is a log-luminosity proxy from absolute $r$-band magnitude, $\ell_r=-0.4M_r$.
   The dashed line is the linear fit on the training split, and color shows the residual $\epsilon_{M|\ell_r}$.
   }
   \label{fig:app_mass_luminosity}
\end{figure}

\begin{figure}[t]
\centering
\begin{minipage}{0.48\linewidth}
\centering
\includegraphics[width=\linewidth]{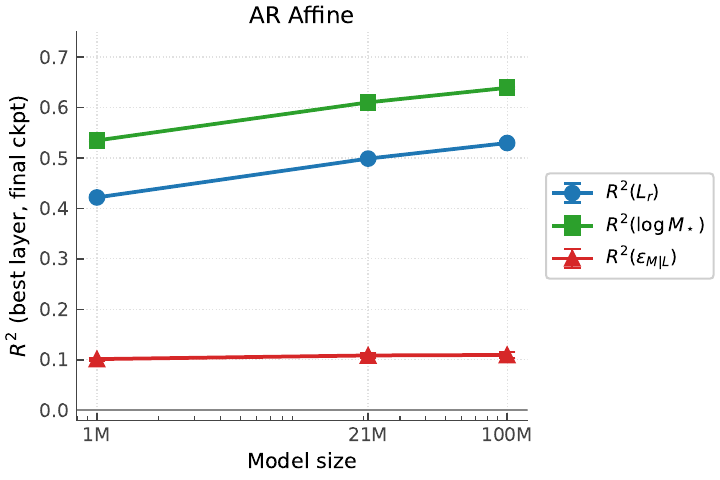}
\end{minipage}
\begin{minipage}{0.48\linewidth}
    \centering
    \includegraphics[width=\linewidth]{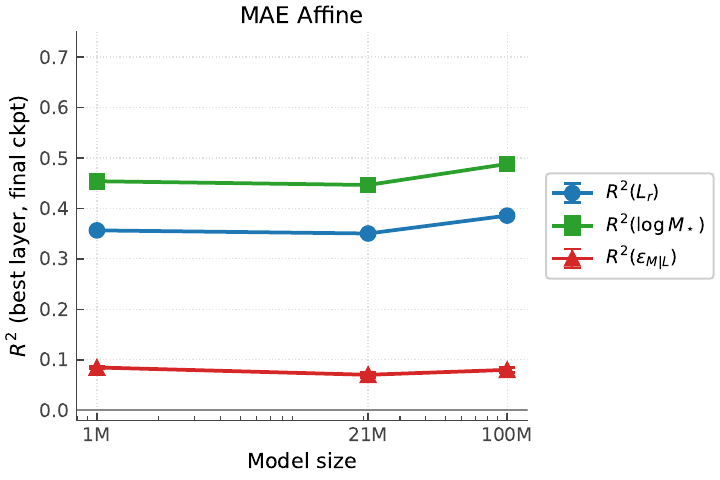}
\end{minipage}

\caption{
\textbf{Residual mass-at-fixed-luminosity remains weak across objective choices.} Each panel reports final-checkpoint, best-layer $R^2$ for the luminosity proxy $\ell_r$, stellar mass $M_\star$, and the residual $\epsilon_{M|\ell_r}$, with error bars showing 95\% bootstrap confidence intervals (300 resamples).
Luminosity and stellar mass are consistently easier to decode, while the residual remains weak across AR/MAE objectives.
}
\label{fig:app_residual_r2_all}

\end{figure}

\begin{figure}[t]
\centering
\begin{minipage}{0.48\linewidth}
\centering
\includegraphics[width=\linewidth]{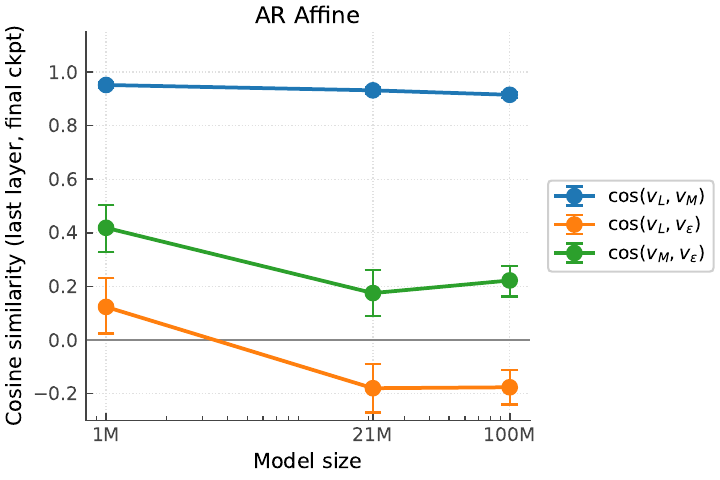}
\end{minipage}
\begin{minipage}{0.48\linewidth}
    \centering
    \includegraphics[width=\linewidth]{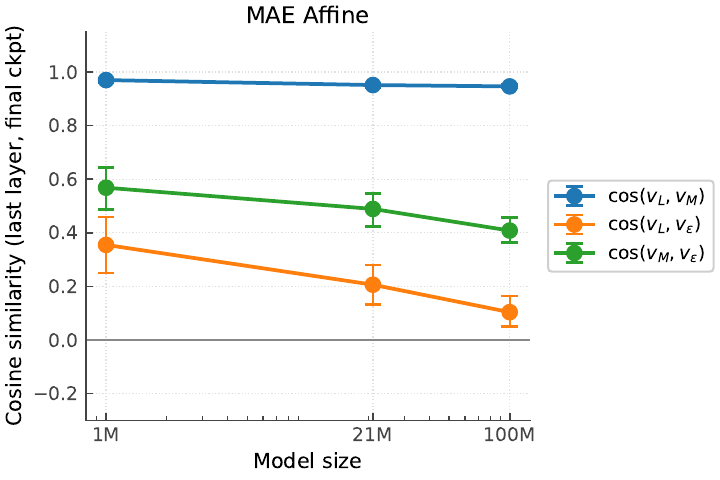}
\end{minipage}

\caption{
\textbf{Residual probe directions are separated from the dominant luminosity-mass axis,} with error bars showing 95\% bootstrap confidence intervals (300 resamples). Across objective choices, luminosity and stellar-mass directions remain strongly aligned. The residual direction is much less aligned with luminosity than the stellar-mass direction is, indicating that the residual probe is not reading out the same luminosity direction. The separation is strongest in the larger AR models and weaker in the MAE setting.
}
\label{fig:app_residual_cosines_all}

\end{figure}

\begin{figure}[t]
   \centering
    \includegraphics[width=\linewidth]{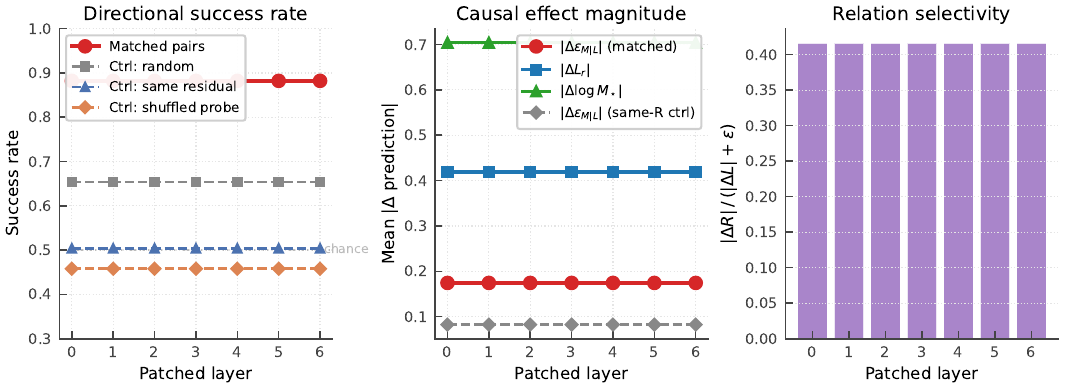}
    \caption{
    Causal tracing of the mass-luminosity residual. 
    Patching activations between galaxy pairs with similar luminosity but different mass residual moves the residual prediction toward the source more often than same-residual and shuffled-probe controls. 
    }
    \label{fig:app_residual_patching}
\end{figure}

Figs.~\ref{fig:app_residual_r2_all} and~\ref{fig:app_residual_cosines_all} show that the residual result is stable across objective choices.
Across configurations, luminosity and stellar mass are decoded much better than the residual $\epsilon_{M|\ell_r}$, which remains weak. 
The cosine plots show that luminosity and stellar-mass directions are strongly aligned, while the residual direction is much less aligned with luminosity than the stellar-mass direction is.
Fig.~\ref{fig:app_residual_patching} then demonstrates an activation-patching check in the AR setting.
We patch between matched galaxy pairs and show that this can move the residual readout above targeted controls. 

\end{document}